\documentclass[10pt,twocolumn,letterpaper]{article}

\usepackage[pagenumbers]{iccv} 

\definecolor{iccvblue}{rgb}{0.21,0.49,0.74}
\usepackage[pagebackref,breaklinks,colorlinks,allcolors=iccvblue]{hyperref}

\def\paperID{11950} 
\def\confName{ICCV}
\def\confYear{2025}

\title{GCRayDiffusion: Pose-Free Surface Reconstruction via \\
Geometric Consistent Ray Diffusion}

\usepackage[perpage]{footmisc}

\def\authorBlock{
    Li-Heng Chen\textsuperscript{1,2} \quad
    Zi-Xin Zou\textsuperscript{2} \quad
    Chang Liu\textsuperscript{1} \quad
    Tianjiao Jing\textsuperscript{1} \quad
    Yan-Pei Cao\textsuperscript{2} \quad
    Shi-Sheng Huang\textsuperscript{1\,\dag} \\[0.5em] 
    Hongbo Fu\textsuperscript{3} \quad
    Hua Huang\textsuperscript{1} \\[0.5em]

    \small
    \textsuperscript{1}Beijing Normal University \quad
    \textsuperscript{2}VAST \quad
    \textsuperscript{3}Hong Kong University of Science and Technology
}

\author{\authorBlock}
\newcommand{\zzx}[1]{{#1}}
\newcommand{\hb}[1]{{         {#1}}}

\begin{document}




\twocolumn[{
\maketitle
\centering
\captionsetup{type=figure}
\includegraphics[width=\textwidth]{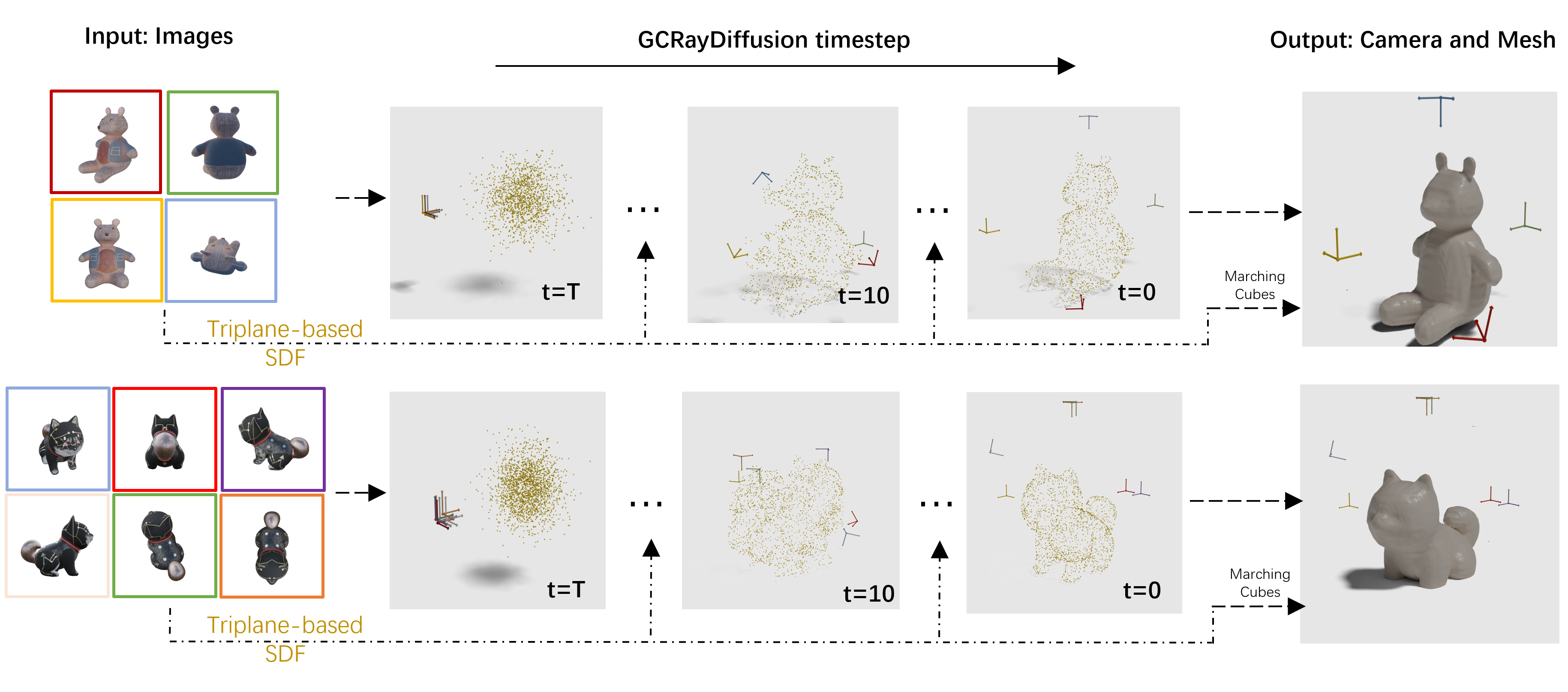}
    \captionof{figure}{\hb{We achieve}
    accurate pose-free neural surface learning with the aid of a novel geometric consistent ray diffusion, i.e., GCRayDiffusion, even from sparse view images (left column). Our GCRayDiffusion model formulates the images' camera poses as neural ray bundles and provides \emph{explicit} sampling points generated during the denoiser processing (middle columns)  to regularize the triplane-based SDF learning, 
    achieving accurate surface reconstruction and camera pose estimation simultaneously (right column). }
\label{fig:teaser}
\vspace{0.5cm}
}]

\def\thefootnote{}\footnotetext{$\dag$ Corresponding authors.}
\begin{abstract}
Accurate surface reconstruction from unposed images \hb{is crucial} 
for efficient 3D object or scene creation\hb{. However, it} 
remains 
challenging \hb{particularly} for the \emph{joint} camera pose estimation. 
 Previous 
approaches have achieved impressive pose-free surface reconstruction results \zzx{in dense-view settings} but could easily fail for sparse-view scenarios without sufficient visual overlap. In this paper, we propose a new \hb{technique for} pose-free surface reconstruction, which follows triplane-based signed distance field (SDF) learning but regularizes the learning by 
\emph{explicit} points sampled from ray-based diffusion of camera pose estimation. 
Our key contribution is a novel Geometric Consistent Ray Diffusion model (GCRayDiffusion), where we represent camera poses as neural bundle rays and regress the distribution of noisy rays via \hb{a} 
diffusion model. 
More importantly, we further condition the \hb{denoising} 
process of RGRayDiffusion using the triplane-based \hb{SDF} 
of the entire scene, which provides effective 3D consistent regularization to get multi-view consistent camera pose estimation. Finally, we incorporate 
RGRayDiffusion to the triplane-based SDF 
learning by introducing on-surface geometric regularization from the sampling points of the neural bundle rays, which leads to highly accurate pose-free surface reconstruction results even for sparse view inputs. Extensive evaluations on public datasets 
show that our GCRayDiffusion 
achieves more accurate camera pose estimation than previous approaches, with \hb{geometrically more} 
consistent surface reconstruction results, 
especially given sparse view inputs.           
\end{abstract}    

\section{Introduction}

3D surface reconstruction from multi-view images has been a long-standing research topic in computer graphics and vision communities. It 
serves as a crucial 3D content creation tool for various applications such as VR/AR, video games, and robotics. We have seen significant progress made from the recent neural surface reconstruction using deep implicit representation~\cite{park2019deepsdf,peng2020convolutional,atzmon2020sal,jiang2020local}, neural radiance field (NeRF)~\cite{mildenhall2021nerf,sun2022neural,yang2022recursive,jiang2022nerffaceediting,somraj2023vip,tancik2023nerfstudio}, and 3D Gaussian Splatting (3DGS)~\cite{Kerbl20233DGS,lyu20243dgsr,kirschstein2024gghead,kerbl2024hierarchical,wu2024recent}. However, most of these 
approaches rely on highly accurate camera pose information as input for each image and would easily fail 
given less accurate camera poses like noisy views \hb{or unknown camera poses}.

For pose-free surface reconstruction, one traditional solution is to first 
estimate the camera poses using the Structure-of-Motion (S\emph{f}M) technique~\cite{snavely2006photo,schonberger2016structure} and then 
perform surface reconstruction according to the estimated camera poses. However, the vanilla S\emph{f}M technique needs dense viewpoints between images with sufficient overlap and 
would cause 
unsatisfactory pose estimation for sparse view images with little visual overlap. For robust pose estimation from sparse view images, subsequent works 
directly regress the camera pose parameters from wide baseline images~\cite{choi2015robust,xiang2018posecnn,balntas2018relocnet,cai2021extreme,rockwell20228}, predict the relative pose probability distributions~\cite{chen2021wide,zhang2022relpose,lin2024relpose++}, or use an iterative refinement strategy~\cite{sinha2023sparsepose}, but the pose estimation quality is still limited. Recent 
works 
represent the camera poses as a joint distribution conditioned on image observations~\cite{wang2023posediffusion} or rays~\cite{zhang2024cameras} and regress the camera poses via the denoiser process of diffusion models, achieving 
impressive camera pose estimation results. However, such diffusion-aided approaches still rely on dense feature matching~\cite{wang2023posediffusion} and fail to 
perform effective bundle adjustment for sparse view scenarios~\cite{zhang2024cameras}.

On the other hand, some recent works propose to jointly learn the neural surface representation and camera poses, leveraging the geometric cues from photometric~\cite{chng2022gaussian,lin2021barf,meng2021gnerf,wang2021nerfmm,yariv2020multiview}, silhouettes~\cite{bosssamurai,kuang2022neroic,zhang2021ners}, or depth points~\cite{wang2024dust3r}. However, those joint learning strategies are only performed independently across dense input images. Some subsequent works~\cite{lin2021barf,meng2021gnerf,truong2022sparf,huang2024sc,wang2023pf,jiang2024few} further explore the extra relations across multiple views to optimize both the neural surface representation and camera poses, achieving more accurate neural surface reconstruction results. However, these approaches still need accurate geometric priors, such as depth~\cite{truong2022sparf} within sufficient overlaps~\cite{huang2024sc} or extra camera intrinsic information~\cite{wang2023pf}. They could not 
guarantee \hb{geometrically} 
consistent surface reconstruction quality given sparse view inputs in many highly freeform applications with unbounded scenarios.   

We propose a new pose-free surface reconstruction approach, which leverages 
effective diffusion-based bundle adjustment to achieve multi-view consistent camera pose estimation and simultaneously leads to geometric consistent surface reconstruction quality even given sparse view inputs. Based on 
a triplane-based SDF 
learning of an entire scene from multiple images, we incorporate 
geometric priors from multi-view consistent camera pose bundle adjustment to regularize the neural implicit field learning. Inspired by the recent ray-based camera parametrization~\cite{zhang2024cameras}, we introduce a new neural bundle ray representation to over-parameterize camera poses but with an extra depth attribution. Leveraging the depth information, we can trace the end points of the neural bundle rays, which can serve as the \emph{explicit} sampling points of the on-surface geometry, thus enabling a differentiable connection between the camera pose and neural implicit field representation. More importantly, we build a Geometric Consistent Ray Diffusion (GCRayDiffusion) model to regress the noisy rays, 
conditioned on the triplane-based SDF 
of the entire scene for multi-view consistent camera pose estimation. Finally, we incorporate the denoiser process of GCRayDiffusion to the triplane-based SDF learning by leveraging the on-surface geometry regularization from the sampling points of the neural bundle rays. Our approach 
leads to multi-view consistent camera pose estimation and geometric consistent surface reconstruction simultaneously, as shown in Fig.~\ref{fig:teaser}. 

To evaluate the effectiveness, we perform extensive evaluations of our approach on publicly released datasets, such as the Objaverse dataset~\cite{objaverse} 
and the Google Scanned Object (GSO)~\cite{downs2022google} dataset, by comparing with state-of-the-art camera pose estimation approaches, including COLMAP~\cite{schonberger2016structure}, RelPose++~\cite{lin2024relpose++}, PoseDiffusion~\cite{wang2023posediffusion}, RayDiffusion \cite{zhang2024cameras} and neural surface reconstruction approaches, such as FORGE~\cite{jiang2024few}, DUSt3R~\cite{wang2024dust3r}. According to the quantitative and qualitative comparisons, our approach 
achieves much better robustness and accuracy in camera pose estimation than those previous approaches, and also geometrically more consistent surface reconstruction results, especially given sparse view image inputs. 

\begin{figure*}
    \centering
    \includegraphics[width=\linewidth]{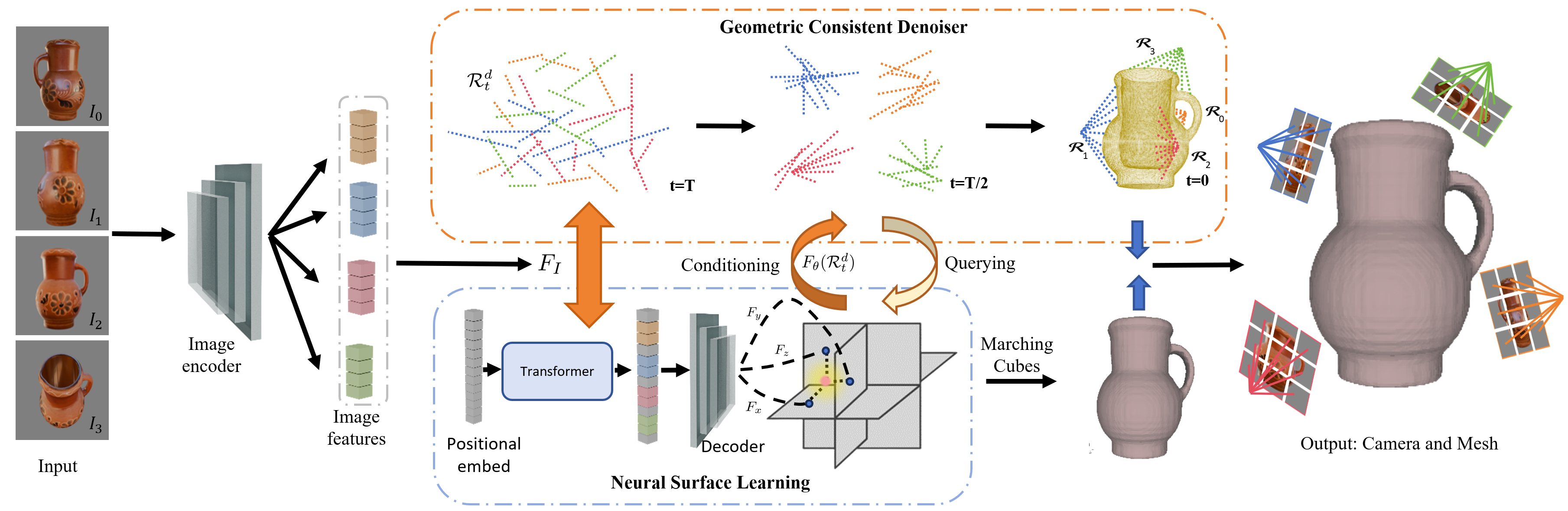}
    \caption{The pipeline of our GCRayDiffusion. Given sparse view images $\mathcal{I}$, our approach extract the image features $F_{\mathcal{I}}$ using an image encoder, and feed $F_{\mathcal{I}}$ to two sub-branches: (1) Geometric Consistent Denoiser processing, which regresses the neural ray bundles $\mathcal{R}^{d}_t$ following a SDF conditioned ray-based diffusion, to estimate the camera poses, and (2) Neural Surface Learning of a triplane-based SDF $ F_{\theta}(\mathcal{R}^{d}_t)$. During the ray bundles denoiser processing, we generate explicit sampling points from neural ray bundles to regularizing the neural surface learning, by querying their SDFs from the triplane-based SDF and locating their position on the surface of the object shape, which leads to accurate surface reconstruction and camera poses estimation simuntaneously. }  
    \label{fig:pipeline}
\end{figure*}


\section{Related Work}
\subsection{Camera Pose Estimation}
The classical Structure-from-Motion (S\emph{f}M)~\cite{snavely2006photo,schonberger2016structure} has been a traditional solution to estimate camera poses from unordered images, which basically relies on finding feature points~\cite{lucas1981iterative} in overlapping images and performs camera poses optimization using Bundle Adjustment~\cite{triggs2000bundle}. Subsequent approaches have significantly improve the SfM quality by improving the feature quality~\cite{detone2018superpoint}, correspondences~\cite{shen2020ransac,yang2019volumetric,sarlin2020superglue} and differentiable Bundle Adjustment~\cite{tang2018ba,lindenberger2021pixel}. However, the SfM framework still rely on dense feature points to estimate camera poses for images with sufficient overlaps, which would lead to significant quality decrease for sparse view images with little overlaps.

To perform camera poses estimation from sparse view images, recent efforts have explored to directly regress 6DoF camera poses from sparse images~\cite{choi2015robust,xiang2018posecnn,balntas2018relocnet,cai2021extreme,rockwell20228}, or predict the probabilistic distribution of relative pose~\cite{zhang2022relpose,lin2024relpose++} using energy-based models. SparsePose~\cite{sinha2023sparsepose} proposed to iteratively refine the sparse camera poses from the initial estimation. RelPose++~\cite{lin2024relpose++} further defines a new camera pose coordinate system and decouples the rotation and translation prediction for more robust camera pose estimation. More recently, with the success of Diffusion models~\cite{10.5555/3495724.3496298,sohl2015deep}, PoseDiffusion~\cite{wang2023posediffusion} proposed to regress the camera pose using a diffusion-aided bundle adjustment. Zhang et al~\cite{zhang2024cameras} introduce bundle rays for even sparse view image inputs. Our approach for camera pose estimation is inspired by these previous approaches, but leverage the geometric prior guidance to the ray based diffusion to achieve multi-view consistent camera pose estimation. 


\subsection{Neural Surface Reconstruction}
There have already been significant progress made for neural surface reconstruction from image sets, by representing scene geometry as deep implicit representation~\cite{park2019deepsdf,peng2020convolutional,atzmon2020sal,jiang2020local}, NeRF~\cite{mildenhall2021nerf,sun2022neural,yang2022recursive,wang2021neus} or 3D Gaussian Splatting~\cite{Kerbl20233DGS,lyu20243dgsr,kirschstein2024gghead,yu2024gaussian}. Subsequent works further incorporate more explicit surface supervisions~\cite{fu2022geo}, surface rendering~\cite{oechsle2021unisurf} or multi-view geometry priors~\cite{darmon2022improving} for more accurate surface learning. However, most of these previous works dense input views for accurate neural surface learning, which would not work for sparse view inputs scenarios. Recently, SparseNeuS~\cite{long2022sparseneus} achieves more generalizable neural surface learning form sparse input views, but still relies on highly accurate camera poses. Unlike these previous neural surface reconstruction works, our approach enables geometric consistent surface reconstruction directly from unposed sparse images, which performs camera pose estimation using a ray based diffusion during the neural surface learning simultaneously. 

\subsection{Joint Implicit Learning and Pose Optimization}
Another category approaches for pose-free surface reconstruction is to jointly perform implicit field learning and camera pose optimization. BARF~\cite{lin2021barf} would be probably one of the first works to the adjust the camera pose directly on NeRF representation following a coarse-to-fine registration strategy. GARF~\cite{chng2022gaussian} further improve the robustness of camera pose refinement using Gaussian based activation functions. SCNeRF~\cite{jeong2021self} proposed to optimize the ray intersection re-projection error during the NeRF learning to adjust camera poses, with subsequent efforts made for more accurate joint learning leveraging more geometric cues such as silhouette~\cite{bosssamurai,kuang2022neroic} or semantic mask~\cite{zhang2021ners}. However, most of the approaches depends on dense input views~\cite{xiao2023level} to perform the joint implicit learning and pose estimation, which will not be effective for sparse scenarios~\cite{chen2021mvsnerf}. 

Recently, for sparse view scenarios, SPARF~\cite{truong2022sparf} proposed to jointly learn the neural surface and refine camera poses using the depth priors. SC-NeuS~\cite{huang2024sc} introduced a joint learning of camera poses and deep implicit representation via the explicit regularization from on-surface geometry. FORGE~\cite{jiang2024few} established cross-view correlations to estimate relative camera poses, which in turn improves the object surface learning. PF-LRM~\cite{wang2023pf} and DUSt3R~\cite{wang2024dust3r} predict sparse poses by predicting pixel-aligned pointclouds and using PnP to recover cameras. 

Inspired by those previous work, our approach proposes to incorporate the explicit regularization from diffusion-based camera pose estimation, to the triplane-based signed distance field learning, which achieves more geometric consistent surface reconstruction results with multi-view consistent camera pose estimation at the same time. 

\begin{figure}
    \centering
    \includegraphics[width=\linewidth]{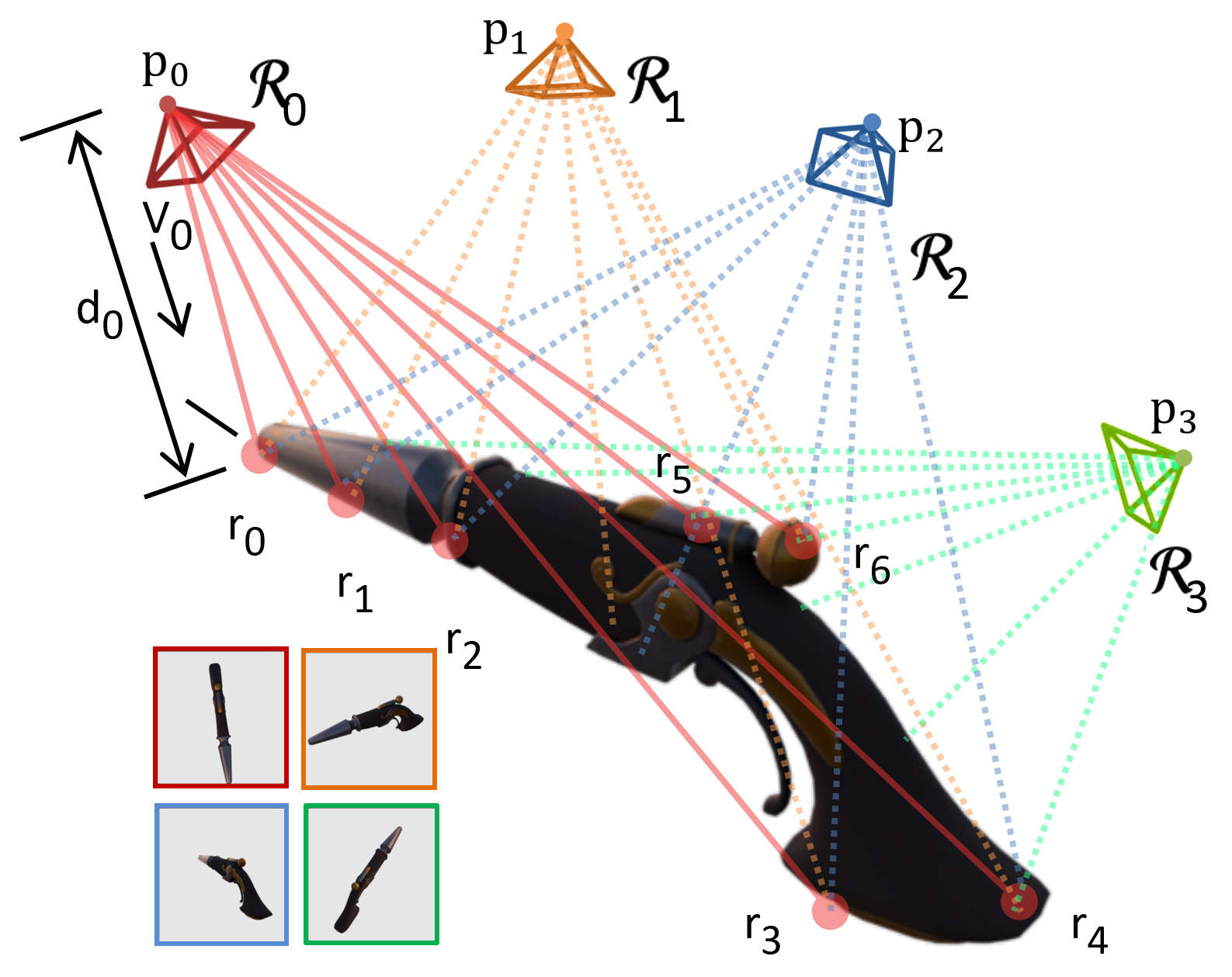}
    \caption{The illustration of our neural bundle rays definition.}
    \label{fig:rays}
\end{figure}

\section{Method}

Given a sparsely sampled image set $\mathcal{I}=\{I_i|i=1,...,N\}$ with $N$ is the image number, our goal is to estimate the camera poses $\mathcal{T}=\{T_i|i=1,...,N\}$ for each image $I_i$, and perform the surface reconstruction represented as neural signed distance field (SDF), i.e., $F_{\theta}(x\in R^3 | \theta) = s \in R$ with $\theta$ represents the parameters of the deep network and $s$ is the signed distance value.  

Inspired by previous approach~\cite{grossberg2001general,schops2020having,zhang2024cameras}, we first propose a new camera pose parametrization using a set of neural bundle rays $\mathcal{R}=\{\textbf{r}_k|k=1,...,M\}$ to parameterize each camera, with the key difference that each ray $\textbf{r}_k$ is additionally affiliated with depth information (Section ~\ref{sec:3.1}). Secondly, we formulate the distribution of noisy rays conditioned on image feature embedding and the signed distance field $F_{\theta}$, and model the camera pose estimation process as a Geometric Consistent Ray Diffusion model (GCRayDiffusion ), to recover camera poses $\mathcal{T}$ by learning the denoiser process of the GCRayDiffusion (Section~\ref{sec:3.2}). Finally, we construct the triplane-based SDF learning by cooperating the on-surface geometry regularization from the sampling points of the neural bundle rays, which can lead to multi-view consistent camera pose estimation and geometric consistent neural surface learning simultaneously (Section~\ref{sec:3.3}). Fig.~\ref{fig:pipeline} illustrates the main pipeline of our approach.  

\subsection{Neural Bundle Ray Representation}
\label{sec:3.1}
Unlike most of the previous approaches that representing camera poses $T_i$ with a 6DoF vector (including 3D rotation and translation), we follow the latest ray-based parametrization introduced by~\cite{zhang2024cameras} for a more flexible camera pose representation. But different from~\cite{zhang2024cameras}, we additionally record the depth information, which indicates the distance from the intersect point that the ray intersect with the shape surface. In this way, we can \emph{explicitly} trace the on-surface sampling points for each ray, thus constructing a differentiable bridge between camera pose and surface representation for thereafter diffusion-aided neural surface learning.

Specifically, as shown in Fig.~\ref{fig:rays}, we propose to over-parameterize each image $I_i$ as a set of neural bundle rays $\mathcal{R}_i = \{\textbf{r}_k^i|k=1,...,M\}$, with each ray $\textbf{r}_k^i$ is represented as a 7-dimension vector including a unit directional vector $\textbf{v}^i_k \in R^3$ though any point $\textbf{p}^i_k \in R^3$ and depth $d^i_k \in R$ following Pl$\ddot{u}$cker coordinates~\cite{Plucker1828} as:
\begin{equation}
    \textbf{r}_k^i = (\textbf{v}^i_k , \textbf{m}^i_k , \textbf{d}^i_k) \in R^7,
\end{equation}
where $\textbf{m}^i_k=\textbf{p}^i_k \times \textbf{v}^i_k \in R^3$ is the moment vector. Given an image $I_i$ with known camera pose, we can uniformly sample a set of 2D pixel coordinates $\{u_k\}_{M}$ to construct the neural bundle rays $\mathcal{R}_i$, and compute the unit directional vector $\textbf{v}_i^k \in R^3$ by unprojecting rays from the pixel coordinates, where the moment vectors $\textbf{m}^i_k$ can be computed by treating the camera centers as the point $p$ since all rays intersect at the camera center. Conversely, given a collection of neural bundle rays $\mathcal{R}_i$ associated with 2D pixels $\{u_k\}_{M}$, we can recover the camera extrinsic and intrinsic by solving the intersection of all rays in $\mathcal{R}_i$. {Please refer to our supplementary materials for the detailed derivations.}

\vspace{0.1cm}
\textbf{Important Property.} Another important difference of our neural bundle ray representation from previous approaches~\cite{zhang2024cameras} is that we can trace the end point $\textbf{r}_d \in R^3$ for each ray $\textbf{r}$, since we record an additional depth information $d$. Specifically, we can compute each ray's end point as $\textbf{r}_d = d \cdot \textbf{v} + \textbf{p}$. The end point $\textbf{r}_d \in R^3$ can also been seen as the intersection point that $\textbf{r}$ intersects with the object's surface, thus connecting the camera pose and object's surface differentiablly, which serves an important property for thereafter ray diffusion aided neural surface learning.

\subsection{Geometric Consistent Ray Diffusion}
\label{sec:3.2}
Based on our above neural bundle ray representation, we view the bundle adjustment process of the noisy rays during the camera pose estimation as a diffusion process, and recover the final rays from initial noisy rays by reversing the Markovian noising process. Specifically, for noisy ray distribution $\mathcal{R}_t \sim q(\mathcal{R}_t), t=0,...,T$, the noising stepsize in the diffusion process is defined by a variance schedule $\{\beta_t\}^{T}_{t=0}$ as:
\begin{equation}
    q(\mathcal{R}_t | \mathcal{R}_{t-1}) = \mathcal{N}(\mathcal{R}_t; \sqrt(1-\beta_t)\mathcal{R}_{t-1}, \beta_t \textbf{I}),
\end{equation}
where $q(\mathcal{R}_t | \mathcal{R}_{t-1})$ is a normal distribution, such that the noisy rays can be computed as:
\begin{equation}
    \mathcal{R}_t = \sqrt{\bar{\alpha}_t} \mathcal{R}_0 + \epsilon \sqrt{1-\bar{\alpha}_t},
\end{equation}
where $\alpha_t = 1-\beta_t$, $\bar{\alpha}_t = 1- \prod^t_{s=0}\alpha_s$, and the noise $\epsilon \sim \mathcal{N}(0,\textbf{I})$.

\vspace{0.1cm}
\textbf{Geometric Consistent Denoiser.} Unlike previous diffusion-based camera pose estimation approaches~\cite{wang2023posediffusion,zhang2024cameras} which directly learn a vanilla denoiser of the target data distribution, we propose to learn a geometric consistent denoiser, which predicts the noise from noisy rays conditioned on the signed distance field (SDF) $F_{\theta}$ of the entire shape. By leveraging such globally consistent geometry prior to the denoiser process, the ray distribution can be effectively bundle-adjusted, thus yielding to multi-view consistent rays prediction for high accurate camera pose estimation. Specifically, as shown in Fig.~\ref{fig:pipeline}, we learn a denoiser network $g_{\phi}$ to predict the noise $\epsilon$ added in the most recent rays $\mathcal{R}_t$ as:
\begin{equation}
    g_{\phi}(\mathcal{R}_t, t | F_{\theta}(\mathcal{R}^{d}_t), F_I ) \to \epsilon,
\end{equation}
where $\mathcal{R}^{d}_t$ is the sampling points set of the rays set $\mathcal{R}_t$, i.e., $\mathcal{R}^{d}_t = \{\textbf{r}_d\}$ with each sampling point $\textbf{r}_d$ is computed from the corresponding ray $\textbf{r}^j \in \mathcal{R}_t$, $F_{\theta}(\mathcal{R}^{d}_t)$ is the signed distance value for the sampling points predicted by $F_{\theta}$, $F_I$ is the feature vector extracted from the original image $I$ at the 2D-pixel coordinate of each ray. To train the denoiser network $g_{\phi}$, we utilize the L2 distance as loss during the parameters optimization following:
\[
L_{diff} = ||g_{\phi}(\mathcal{R}_t, t | F_{\theta}(\mathcal{R}^{d}_t), F_I ) - \epsilon||_{2}.
\]

In this way, we build up a geometric consistent Ray Diffusion (GCRayDiffusion) model to predict the neural bundle rays set of each input image, and recover the corresponding camera pose via the transformation of the neural bundle rays. 

\subsection{Diffusion-aided Neural Surface Learning}
\label{sec:3.3}
Finally, we incorporate the GCRayDiffusion model the the neural surface learning for the surface reconstruction. Our key observation is to leverage the sampling points of the neural bundle rays as explicit regularization to guide the  neural signed distance field (SDF) $F_{\theta}$ learning, thus introducing the multi-view consistent camera pose bundle adjustment priors via GCRayDiffusion for the $F_{\theta}$ learning, towards geometric consistent surface reconstruction results. 

Specifically, as shown in Fig.~\ref{fig:pipeline}, we formulate $F_{\theta}$ as a triplane-based signed distance field (SDF), which consists a Transformer-based image encoder ${\varPhi}$ to extract triplane feature maps from image inputs and a MLP-based decoder $\mathcal{D}$ to regress the SDF prediction, as:
\begin{equation}
    F_{\theta}(x \in R^3 | {\varPhi} , \mathcal{D} ) \to s \in R,
\end{equation}
\[
s.t. \quad {\varPhi}(\mathcal{I}) = \{F_x, F_y, F_z\}, \quad \mathcal{D}(x \in R^3 |F_x, F_y, F_z ) =s,
\]
where $\{F_x, F_y, F_z\}$ are the triplane feature maps and $s$ is the SDF value.

\vspace{0.1cm}
\textbf{Diffusion-aided Learning.} We leverage the sampling points $\mathcal{R}^{d}$ of the neural bundle rays $\mathcal{R}$ during the $T$ step denoiser process of the GCRayDiffusion model, to guide $F_{\theta}(x \in R^3 | {\varPhi} , \mathcal{D} ) $ learning. One straightforward yet effective operation is to regularize such sampling points $\mathcal{R}^{d}_t$ located on the latent geometry surface of $F_{\theta}(x \in R^3 | {\varPhi} , \mathcal{D} )$, i.e., 
\begin{equation}
    F_{\theta}(\mathcal{R}^{d}_t | {\varPhi} , \mathcal{D} ) \to 0,
\end{equation}
during each time step $t$ of the denoiser process from the GCRayDiffusion model. So after $T$ step denoising process, we effectively guide the neural SDF learning using the multi-view consistent camera pose bundle adjustment, which leads to more better geometric consistent surface learning. Once  $F_{\theta}(\mathcal{R}^{d}_t | {\varPhi}, \mathcal{D} )$ is learned, we extract the surface results of the zero level-set of $F_{\theta}(\mathcal{R}^{d}_t | {\varPhi}, \mathcal{D} )$ using Marching Cubes~\cite{lorensen1998marching}. Simultaneously, the final neural bundle rays $\mathcal{R}$ are recovered by the GCRayDiffusion model, which is used to compute the final camera poses $\mathcal{T}$.   

\vspace{0.5cm}
\section{Experiments}
\subsection{Experimental Setup}
\vspace{0.1cm}
\textbf{Dataset and Metrics.} 
{We evaluate our approach on the Objaverse dataset~\cite{objaverse} 
and the Google Scanned Object (GSO)~\cite{downs2022google} dataset.} 
{The Objaverse dataset consists of diverse 3D scenes, while the 
GSO dataset includes 300 samples from unseen, allowing us to test the generalization ability of our method.} 
{To evaluate the accuracy for camera pose estimation, we adopt two accuracy metrics including camera rotation accuracy (within 15 degrees), camera translation accuracy (within 0.1). For the surface reconstruction accuracy, we adopt the following metrics including Hausdorff Distance (HD), Chamfer Distance (CD), Normal Consistency (NC), by measuring the surface mesh extracted from our GCRayDiffusion and the ground truth surface mesh, and also F-score (based on the Hausdorff Distance accuracy) where we use HD threshold as 5\% when calculating the Precision and Recall respectively.}

\vspace{0.1cm}
\textbf{Training Details.} 
{Our training process involves initializing camera poses, for the Objaverse dataset, we utilize random initialization.} 
{The training consists of 40K iterations, with 25K allocated to the coarse stage and 15K to the fine stage.} 
{The weights for the loss functions are set as follows: $w_A=0.1$, $w_{\rho}=w_M=0.001$, and $\lambda_{np}=0.05$.} 
{The local weights are $\lambda_{ml}=\lambda_{nl}=0.01$, and the flatten weight is $\lambda_f=20$.} 
{All experiments are conducted on a single NVIDIA A800 GPU.}

\vspace{0.1cm}
\textbf{Comparing Approaches.} We compare our method with state-of-the-art camera pose estimation approaches, including  RelPose++~\cite{lin2024relpose++}, PoseDiffusion~\cite{wang2023posediffusion}, RayDiffusion~\cite{zhang2024cameras}, FORGE~\cite{jiang2024few} and DUSt3R~\cite{wang2024dust3r}. Besides, we also choose COLMAP~\cite{schonberger2016structure} as baseline approach during the evaluation. During the experiment, we evaluate both the camera pose estimation and surface reconstruction  accuracy. For the camera pose, we directly compute the accuracy metrics (both rotation and translation) for these comparing approaches. As for the surface reconstruction, since COLMAP~\cite{schonberger2016structure}, RelPose++~\cite{lin2024relpose++}, PoseDiffusion~\cite{wang2023posediffusion} and RayDiffusion~\cite{zhang2024cameras} only compute camera poses but didn't perform surface reconstruction. For a fair comparison, we further conduct NeRF-based surface reconstruction using their camera poses estimation. Then we perform the mesh surface quality extracted from all of these comparing approaches to conduct the comparison. We use the public release source code of COLMAP\footnote{https://github.com/colmap/colmap}, RelPose++\footnote{https://github.com/amyxlase/relpose-plus-plus}, PoseDiffusion\footnote{https://github.com/facebookresearch/PoseDiffusion}, RayDiffusion\footnote{https://github.com/jasonyzhang/RayDiffusion}, FORGE\footnote{https://github.com/UT-Austin-RPL/FORGE} and DUSt3R\footnote{https://github.com/naver/dust3r} by using the default parameter configuration for fair comparison. To achieve better performance for COLMAP, we use SuperPoint features~\cite{detone2018superpoint} and SuperGlue matching~\cite{sarlin2020superglue} during the experiments.

\begin{table}[h!]
\centering
\small 
\setlength{\tabcolsep}{5pt} 
\begin{tabular}{lcccccc}
\toprule
 & \multicolumn{5}{c}{\textbf{Rotation Accuracy}} \\ 
\# of images
& 2 & 3 & 4 & 5 & 6 \\ \midrule
COLMAP & 31.20 & 30.16 & 28.74 & 29.89 & 30.69 \\ 
RelPose++ & 61.35 & 62.71 & 65.79 & 66.11 & 68.42 \\ 
FORGE & 89.26 & 89.89 & 88.36 & 79.23 & 78.65 \\
PoseDiffusion & 77.3 & 74.82 & 75.25 & 69.34 & 62.1 \\ 
RayDiffusion & 86.00 & 85.00 & 87.20 & 80.33 & 79.39 \\ 
DUSt3R & 90.52 & 91.87 & 92.26 & 91.58 & 91.16 \\ 
Ours & \textbf{93.21} & \textbf{93.17} & \textbf{92.32} & \textbf{93.60} & \textbf{92.92} \\ 
\midrule
 & \multicolumn{5}{c}{\textbf{Translation Accuracy}} \\ 
 \# of images
& 2 & 3 & 4 & 5 & 6 \\ \midrule
COLMAP & 29.36 & 26.25 & 21.83 & 23.79 & 25.21 \\ 
RelPose++ & 63.24 & 60.55 & 57.31 & 58.12 & 57.46 \\ 
FORGE & 48.54 & 44.39 & 41.33 & 43.58 & 43.26 \\ 
PoseDiffusion & 40.21 & 39.33 & 38.23 & 31.25 & 30.88 \\ 
RayDiffusion & 65.32 & 50.43 & 41.28 & 39.91 & 39.80 \\ 
DUSt3R & 68.26 & 62.45 & 62.03 & 62.97 & 60.21 \\ 
Ours & \textbf{69.77} & \textbf{63.44} & \textbf{62.62} & \textbf{63.91} & \textbf{62.89} \\ 
\bottomrule
\end{tabular}
\caption{\textbf{The camera pose estimation accuracy evaluated on Objaverse dataset.}}
\label{tab:combined_accuracy}
\end{table}

\vspace{0.2cm}
\subsection{Evaluation on Objaverse Dataset}

{We first conduct evaluation on Objaverse dataset, by comparing our approaches with those previous approaches. Since the cases in Objaverse dataset have different number of input images, for a comprehensive evaluation, we conduct experiments by changing the number of input images, i.e., from 2-6 images, for all of the different comparing approaches. 
}

\vspace{0.1cm}
\textbf{Camera Pose Estimation Comparison.} 
{As shown in Table \ref{tab:combined_accuracy}, in terms of camera rotation accuracy and camera translation accuracy, our approach can achieve consistently better accuracy than all of those previous approaches, where our approach significantly outperforms COLMAP, RelPose++ and FORGE respectively, and also better camera pose estimation in rotation and translation accuracy than SOTA approaches such as PoseDiffusion, RayDiffusion and DUSt3R respectively. Besides, given different number of image input (from 2 to 6), our approach also consistently outperform those previous approaches. 

\textbf{Surface Reconstruction Comparison.}
{Except from the camera pose estimation, we also conduct comparison on the surface reconstruction accuracy. As shown in Table~\ref{tab:combined} (upper rows), our approach also achieves consistently better accuracy metrics, including CD, HD, NC and F-scores, than all of those previous approaches. 

\textbf{Qualitative Comparison.}
{The qualitative results presented in Fig. \ref{fig:1} illustrate the high-quality surface reconstructions achieved by our method from Objaverse dataset, and also some of those previous SOTA approaches such as RelPose++ (with NeRF reconstruction), FORGE and DUSt3D respectively. As we can see in the figure, RelPose++ often crush to achieve a complete surface reconstruction, though the camera poses estimation are reasonable, but NeRF fails to conduct success surface reconstruction given such sparse image input. Although FORGE and DUSt3R can achieve reasonable surface reconstruction results, but our approach can achieve accurate surface reconstruction with more geometric details, with the benefit of more accurate camera pose estimation. } 

\begin{figure*}[t]
    \centering
    \includegraphics[width=\linewidth]{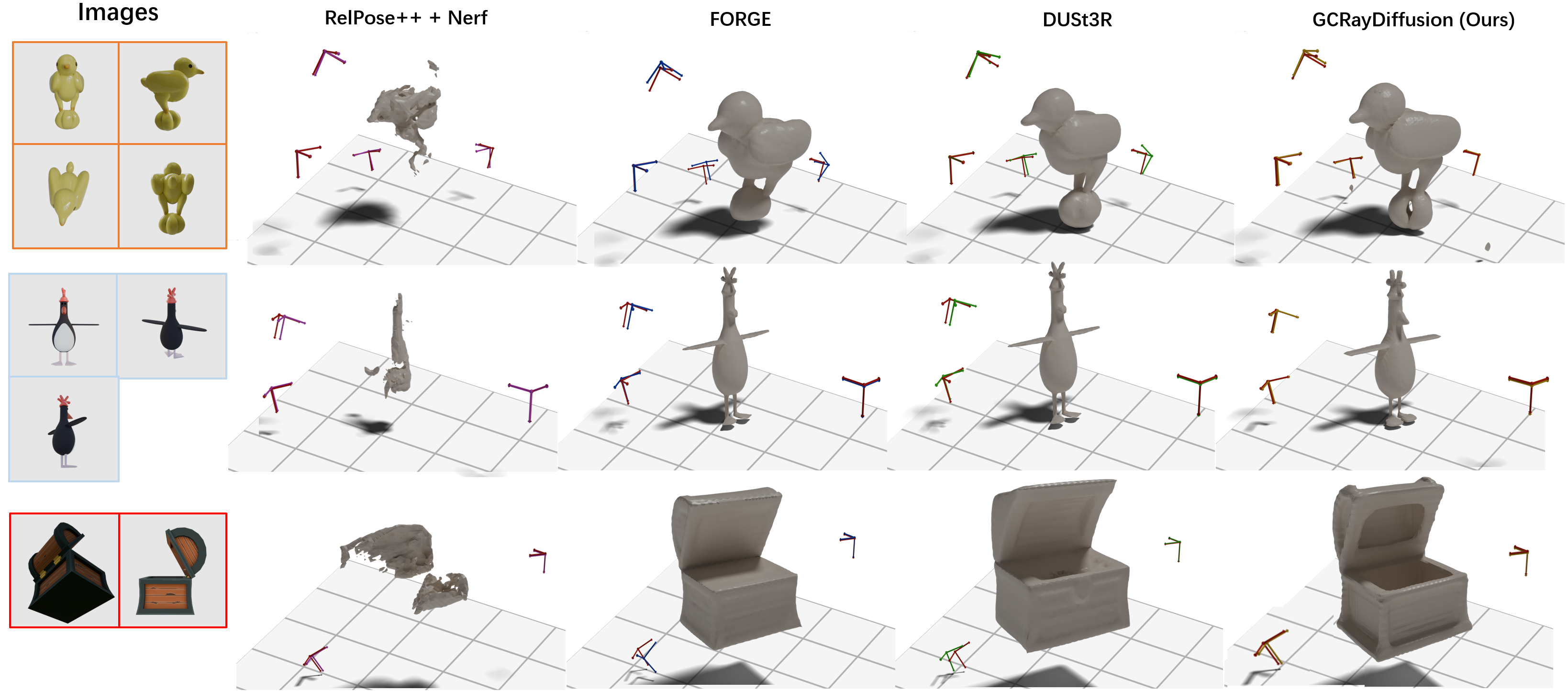}
    \caption{Qualitative surface reconstruction comparison evaluated on Objaverse dataset for different comparing approaches, including RelPose++, FORGE, DUSt3R and our GCRayDiffusion (from left to right column) respectively.  }  
    \label{fig:1}
\end{figure*}

\vspace{0.1cm}
\subsection{Generalization Evaluation to GSO Dataset}

{We also evaluate the generalization ability of our approach to another GSO dataset, where we use the parameter weights pre-trained on Objaverse dataset and conduct test on GSO dataset. Besides, we also make comparison with those previous approaches mentioned above. 
}

\textbf{Camera Pose Estimation Comparison.} 
{As shown in Table \ref{tab:gso_combined_accuracy}, in terms of camera rotation accuracy and camera translation accuracy, our approach can also achieve much better accuracy than all of those previous approaches, where we only achieve slightly worse camera rotation accuracy than DUSt3R when given 4 input images. This means that our approach also achieve consistently better than all of those previous approaches evaluated on GSO dataset. 
} 

\textbf{Surface Reconstruction Comparison.}
{Except from the camera pose estimation, we also conduct comparison on the surface reconstruction accuracy. As shown in Table~\ref{tab:combined} (bottom rows), our approach also achieves consistently better accuracy metrics, including CD, HD, NC and F-scores, than all of those previous approaches, where we only achieve a slightly worse F-score accuracy than DUSt3R. 

\textbf{Qualitative Comparison.}
{The qualitative results presented in Fig. \ref{fig:2} illustrate the high-quality surface reconstructions achieved by our method from the GSO dataset, and also some of those previous SOTA approaches such as RelPose++ (with NeRF reconstruction), FORGE, and DUSt3D respectively. Similarly, our approach can also achieve better visual reconstruction results than those three SOTA approaches. } 

\begin{figure*}
    \centering
    \includegraphics[width=\linewidth]{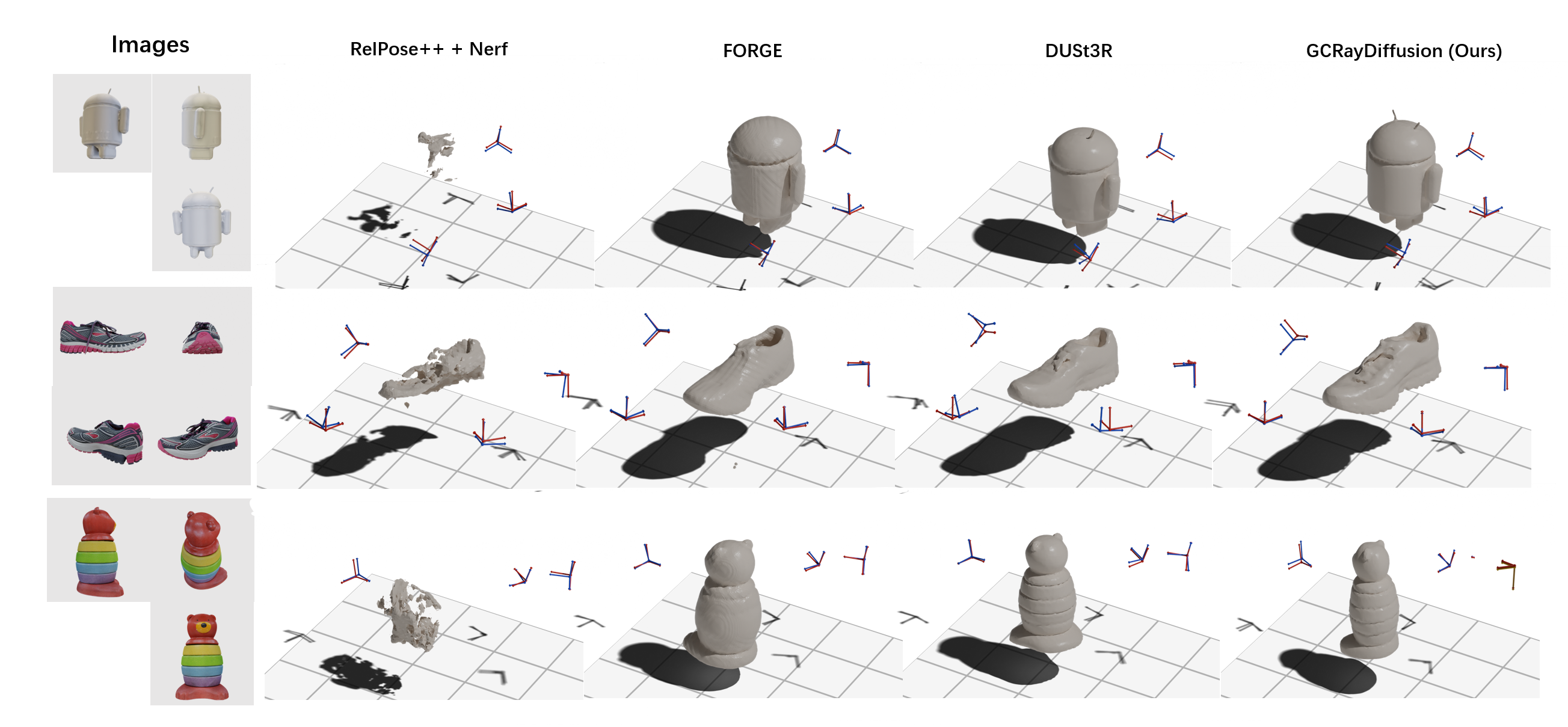}
    \caption{Qualitative surface reconstruction comparison evaluated on GSO dataset for different comparing approaches, including RelPose++, FORGE, DUSt3R, and our GCRayDiffusion (from left to right column) respectively.  }  
    \label{fig:2}
\end{figure*}

\begin{table}[t]
\centering
\small 
\setlength{\tabcolsep}{5pt} 
\begin{tabular}{lcccccc}
\toprule
\# of images & \multicolumn{5}{c}{\textbf{Rotation Accuracy}} \\ 
& 2 & 3 & 4 & 5 & 6 \\ \midrule
COLMAP & 29.23 & 29.58 & 31.45 & 32.15 & 32.50 \\ 
RelPose++ & 59.23 & 59.88 & 62.49 & 64.12 & 66.92 \\ 
FORGE & 83.21 & 84.37 & 85.96 & 79.83 & 76.11 \\ 
PoseDiffusion & 75.48 & 74.99 & 73.31 & 70.08 & 61.25 \\ 
RayDiffusion & 86.33 & 84.89 & 87.31 & 81.22 & 76.3\\ 
DUSt3R & 91.33 & 91.27 & \textbf{92.37} & 90.06 & 91.03 \\ 
Ours & \textbf{93.20} & \textbf{93.23} & 91.35 & \textbf{91.03} & \textbf{94.32} \\ 
\midrule
\# of images & \multicolumn{5}{c}{\textbf{Translation Accuracy}} \\ 
& 2 & 3 & 4 & 5 & 6 \\ \midrule
COLMAP & 26.54 & 23.18 & 21.83 & 22.47 & 20.16 \\ 
RelPose++ & 65.33 & 62.29 & 60.36  & 61.25 & 64.31 \\ 
FORGE & 49.23 & 48.56 & 45.88 & 46.21 & 42.19 \\
PoseDiffusion & 41.33 & 38.79 & 39.31 & 34.27 & 29.06 \\ 
RayDiffusion & 63.42 & 50.25 & 41.79 & 38.41 & 38.02 \\ 
DUSt3R & 66.33 & 61.9 & 61.93 & 59.82 & 60.59 \\ 
Ours & \textbf{68.82} & \textbf{62.77} & \textbf{61.95} & \textbf{63.13} & \textbf{64.81} \\ 
\bottomrule
\end{tabular}
\caption{\textbf{The camera pose estimation accuracy evaluated on the GSO dataset.}}
\label{tab:gso_combined_accuracy}
\end{table}

\begin{table}[h!]
\centering
\small 
\setlength{\tabcolsep}{5pt} 
\begin{tabular}{lccccc}
\toprule
\textbf{Dataset} & \textbf{CD}$\downarrow$ & \textbf{HD}$\downarrow$ & \textbf{NC}$\uparrow$ & \textbf{F-score}$\uparrow$ \\ \midrule
\textbf{Objaverse} & & & & \\
\quad COLMAP & 9.17 & 15.41 & 0.74 & 0.536 \\
\quad RelPose++ & 4.58 & 6.49 & 0.76 & 0.615 \\
\quad FORGE & 0.145 & 0.405 & 0.989 & 0.89 \\
\quad DUSt3R & 0.132 & 0.368 & 0.995 & 0.97 \\
\quad Ours & \textbf{0.125} & \textbf{0.323} & \textbf{0.997} & \textbf{0.99} \\ \midrule
\textbf{GSO} & & & & \\
\quad COLMAP  & 10.26 & 13.32 & 0.72 & 0.519 \\
\quad RelPose++ & 3.96 & 5.72 & 0.73 & 0.527 \\
\quad FORGE & 0.155 & 0.437 & 0.985 & 0.854 \\
\quad DUSt3R & 0.139 & 0.366 & 0.973 & \textbf{0.961} \\
\quad Ours & \textbf{0.131} & \textbf{0.302} & \textbf{0.988} & 0.958 \\ \bottomrule
\end{tabular}
\caption{\textbf{Surface reconstruction accuracy on Objaverse and GSO dataset respectively.}}
\label{tab:combined}
\end{table}

\subsection{Ablation}

\begin{figure}
    \centering
    \includegraphics[width=1.05\linewidth]{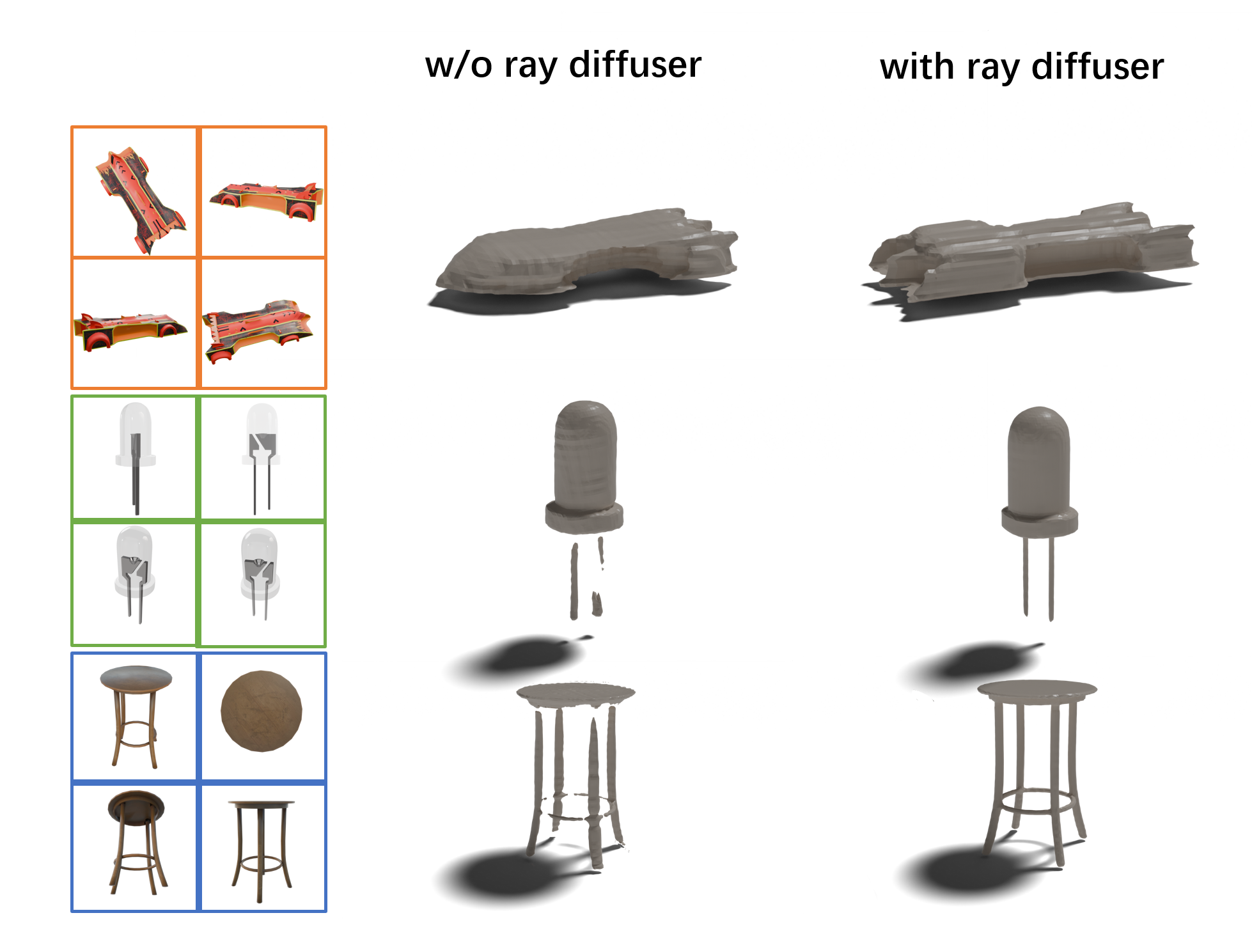}
    \caption{Surface reconstruction comparison with or without using ray diffuser of our GCRayDiffusion.  }  
    \label{fig:Ablation}
\end{figure}

We designed an ablation experiment to study how the two main components impact the final camera pose estimation and surface reconstruction respectively, including (1) how the ray diffusion performs without the condition of triplane-based SDF (termed as 'w/o SDF') for the camera pose estimation, and (2) how the triplane-based SDF learning perform without the aid of our GCRayDiffusion (termed as 'w/o ray diffuser') for the surface reconstruction. As shown in Table~\ref{tab:combined_accuracy1} and Table~\ref{tab:combined2} evaluated on the Objaverse dataset, we can see that both the camera pose estimation and surface reconstruction quality will decrease without using the two main components. Fig.~\ref{fig:Ablation} also show several surface reconstruction results with or without using the guide of our GCRayDiffusion respectively. 


\begin{table}[h!]
\centering
\begin{tabular}{ccc}
\toprule
  & \textbf{Rotation} & \textbf{Trans} \\ \midrule
w/o SDF & 86.3 & 37.5 \\
GCRaydiffusion &  92.32 & 62.62\\
\midrule
\end{tabular}
\caption{\textbf{Camera pose estimation accuracy comparison.}}
\label{tab:combined_accuracy1}
\end{table}

\begin{table}[h!]
\centering
\begin{tabular}{lccccc}
\toprule
 & \textbf{CD}$\downarrow$ & \textbf{HD}$\downarrow$ & \textbf{NC}$\uparrow$ & \textbf{F-score}$\uparrow$ \\ \midrule
 w/o ray diffuser & 0.16 & 0.802 & 0.992 & 0.988\\ 
 GCRaydiffusion & 0.125 & 0.323 & 0.997 & 0.99 \\\midrule
\end{tabular}
\caption{\textbf{Surface reconstruction accuracy comparison.}}
\label{tab:combined2}
\end{table}

\section{Conclusion}
This paper contributes a new pose-free surface learning with the aid of a novel geometric consistent ray diffusion, i.e., GCRayDiffusion, which achieves better camera pose estimation and surface reconstruction than previous SOTA approaches. We hope that our approach can inspire subsequent works for more robust and accurate pose-free surface reconstruction from sparse image inputs in this community.

{
    \small
    \bibliographystyle{ieeenat_fullname}
    \bibliography{reference}
}

\end{document}